\documentclass{article}
\usepackage[utf8]{inputenc}
\usepackage[margin=0.75in]{geometry}
\usepackage{authblk}
\usepackage[T1]{fontenc}
\usepackage[utf8]{inputenc}
\usepackage{changepage}
\usepackage{multicol}
\usepackage{blindtext}
\usepackage{floatrow}
\usepackage{caption}
\hyphenchar\font=-1
\title{A smartphone based multi input workflow for non-invasive estimation of haemoglobin levels using machine learning techniques}
\author[2]{Sarah}
\author[2]{S{.} Sidhartha Narayan}
\author[2]{Irfaan Arif}
\author[2]{Hrithwik Shalu}
\author[1]{Juned Kadiwala}

\affil[1]{University of Cambridge}
\affil[2]{Indian Institute of Technology, Madras}

\date{}

\usepackage{natbib}
\usepackage{graphicx}

\begin{document}

\maketitle

\begin{adjustwidth*}{+2cm}{+2cm}
\begin{center}
   \emph{Abstract}
\end{center}

   \emph{We suggest a low-cost, non-invasive healthcare system that measures haemoglobin levels in patients and can be used as a preliminary diagnostic test for anaemia. A combination of image processing, machine learning and deep learning techniques are employed to develop predictive models to measure haemoglobin levels. This is achieved through the color analysis of the patient’s fingernail beds, palpebral conjunctiva and tongue. This predictive model is then encapsulated in a healthcare application. This application expedites data collection and facilitates active learning of the model. It also incorporates personalized calibration of the model for each patient, assisting in the continual monitoring of the haemoglobin levels of the patient. Upon validating this framework using data, it can serve as a highly accurate preliminary diagnostic test for anaemia.}
\end{adjustwidth*}

\section{Introduction}
\begin{multicols}{2}

Anaemia is a condition in which the number of red blood cells, and consequently their oxygen-carrying capacity is insufficient to meet the body’s physiological needs.\textsuperscript{1} It is a serious global public health problem, affecting 1.62 billion people (24.8\% of the total population) worldwide. The groups most vulnerable to this disease are children under the age of five (47.4\%) and pregnant women (41.8\%).\textsuperscript{2}

Currently, the gold standard for diagnosing anaemia is the Complete Blood Count (CBC) test. CBC measures the Haemoglobin (Hb) levels, the Hematocrit (HCT) levels which is a measure of how much space red blood cells take up in a patient’s blood and finally the Mean Corpuscular Volume (MCV) which is the average size of a patient’s red blood cells, cluing further into the cause of anaemia. A CBC test requires blood sampling by a trained phlebotomist, a clinical haematology analyzer with the required electrical power, biochemical reagents, and infrastructure thereof, along with a trained laboratory technician to perform the analysis\textsuperscript.{3} It is invasive and time-consuming, rendering it futile in clinical circumstances where an immediate knowledge of a patient’s Hb levels could have led to enhanced patient care.

The development of a non-invasive, low-cost and rapid-detection technique would mean Hb levels can be taken alongside vital signs when a patient is admitted. It would prove especially powerful in countries lacking in healthcare infrastructure. The Hb levels in blood changes slowly. Such a technique would also allow for continuous monitoring of Hb levels as a result of frequent testing.\textsuperscript{8}

Several reports have shown that Hb levels in blood positively correlate with the pallor of various anatomic regions of a patient’s body, namely fingernail beds, palpebral conjunctiva, tongue and palmar creases.\textsuperscript{4,5,6,7} In a controlled lighting environment, there is a good relationship between the red pixels of the conjunctiva and Hb levels.\textsuperscript{10} Another study found the difference in mean red pixel intensities and green pixel intensities are smaller for the conjunctiva images of the anaemic patients when compared to non-anaemic ones.\textsuperscript{9} Similar approaches correlate the Erythema Index, the G component of RGB colour space and the a* \& b* component of CIELab colour space with Hb values.\textsuperscript{11,12,13}

A major issue with these image processing methods is that poorly taken pictures can undermine the whole process. One counter-approach is using standard colorimetry practices that require a reference element to be captured together with the subject to extract reliable colour values from the digital image. In one study, the white sclera of the eye was extracted from the image taken and used a reference of the brightness of the image. The same study extracted twelve colour planes across four colour spaces, together with Frangi-filtering and gradient filtering, enhancing the image separability for pallor severity in eye and tongue pallor images and giving 27 features being passed to auto-parameterised data-models.\textsuperscript{7}

Hb level estimation can also be done with the image of fingernail beds, by including the image metadata along with the color plane data and using a robust multi-linear regression algorithm. This approach also allows for personalised calibration and effortless real-time monitoring of Hb levels. This same study also shows a positive correlation of pixel values with HCT levels of the patient. The MCV levels do not show any such definite relationship.\textsuperscript{6}

In the present study, data models created use images of fingernail beds, conjunctiva and tongue, the pallor information of the three regions cross validating one another to predict the Hb levels of a patient more accurately and determine the type of anaemia the patient is suffering from. We employ several combinations of classification and regression algorithms and determine the best possible combination.

   \end{multicols}
   
\section{Concepts}
\begin{multicols}{2}
Fingernail beads and Palpebral Conjunctiva do not contain melanin producing skin cells. The primary color is blood haemoglobin. Hence, their pallor seldom varies with the skin colour of the subject. For our study, we have chosen fingernail beds, palpebral conjunctiva and tongue for colour profile analysis. The images of these regions can easily be captured through a smart phone and without the need for any other equipment.

\emph{Palpebral Conjunctiva}: Palpebral Conjunctiva is the clear, thin membrane covering the inner surface of both the upper and lower eyelids. This region has many small blood vessels to provide nutrients to the eye and lids. The abundant presence of microvessels in the region makes it a perfect candidate to carry out non-invasive and highly precise imaging studies.

\emph{Fingernail beds}: Fingernails are the easiest for a user to self-image. They also have a low person-to-person size and shape variability, thus contributing to the practicality of our assessments.\textsuperscript{6}

\end{multicols}

\begin{figure}[h!]
\centering
\includegraphics[scale=0.45]{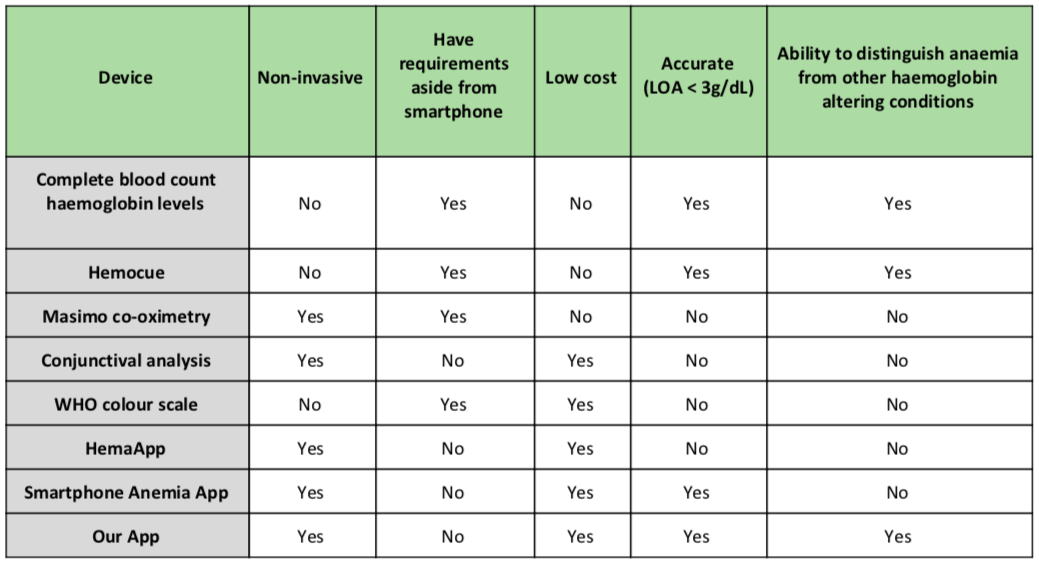}
\caption*{Table 1 : Comparison of existing anaemia diagnostic technologies}
\end{figure}

\begin{multicols}{2}
\emph{Tongue}: A number of clinical indices contribute to the tongue colour14, making it imprecise to solely attribute its pallor to anaemic causes. However, one can use the information on tongue pallor in combination with conjunctiva and fingernails to improve model accuracy and spot any abnormalities. Certain characteristic colours exhibited by the tongue can also be correlated with the cause or subtype of anaemia.

Physicians rely on a qualitative clinical assessment of the conjunctival pallor, tongue pallor and pallor of the patient’s nail beds to gauge the severity of a patient’s anaemia. This preliminary assessment is followed by a complete blood count test for accurately measuring haemoglobin levels. Current technologies in digital image processing and artificial intelligence, can enhance the qualitative preliminary assessment in this case, leading us to quantitatively correlate the severity of a patient’s anaemia and their clinical pallor. This could also potentially eliminate the need of complete blood count tests in some cases.

Table 1 enumerates and compares existing diagnostic methods of anaemia. We have surveyed each method on its invasiveness, requirements, price, accuracy and its ability to provide additional information about the sub-type of anaemia present.

\end{multicols}
\section{Proposed Methodology}
\begin{multicols}{2}
\noindent\emph{A. 	Input Images} \\

In imaging studies, it is crucial to ensure consistent lighting conditions, while taking images. Since it is not always a possibility, we have incorporated illumination correction algorithms while processing images. Also, we need to avoid camera flashes and inconsistent illumination of the regions of interest. Improper contrast, glare or flicker while taking images can compromise the accuracy or reproducibility of the results.

It is important to ensure that the conjunctiva is pulled down sufficiently for capturing the maximum area possible. All the fingernail beds need to be captured from both hands to average out inconsistencies while training the model. The tongue should be properly cleaned too. The image of the blood report of the patient is also captured. A cloud-based natural language processing algorithm automatically reads the patient’s Haemoglobin levels from the image of the blood report, and stores it in our database alongside the captured images.\\

\noindent\emph{B.	Data Augmentation} \\

Artificially increasing the number of training samples improves the derived model. We flip, rotate and/or affine shift the images to increase the instances of each class in the dataset. We refrain from using blurring or diffusion-based methods of data augmentation, as this would distort pixel intensities and interfere with the results.

We are bound to have an imbalance in the dataset, either having more anaemic or more non-anaemic patient data. To combat this, we use ROSE (Random Oversampling) and SMOTE (Synthetic Minority Oversampling Technique), which will balance different classes of data.\\

\noindent\emph{C.	Image Preprocessing}\\

For pre-processing of images, we rely on a variant of Adaptive Histogram Equalisation (AHE). This method computes several histograms corresponding to distinct sections of the images, and uses them to redistribute the light in the image. It is suitable for improving local contrast and enhancing edge definitions, which would later help in improved segmentation of the regions of interest in the image. However, AHE by itself amplifies the noise, we instead use a version of it called CLAHE (Contrast Limited Adaptive Histogram Equalisation). CLAHE discards the part of the histogram that exceeds a clip limit and redistributes it equally among all histogram bins.

The equalisation is performed on the Y intensity plane of the image converted to the YCbCr colour space, followed by adaptive thresholding. We have also applied morphological transformation on the images, assisting us to remove regions in the image distorted by noise and texture.

For palpebral conjunctiva, while segmenting the conjunctiva, we extract the sclera of the eye and use it as a soft reference frame for illumination correction in the rest of the image. A CRF (Conditional Random Field) enhances and solidifies the generated segmentation masks. A CRF improves prediction by taking into account the labels of neighbouring samples and building a context for the current sample.\\

\noindent\emph{D.	Segmentation}\\

We have used two major techniques for segmenting our regions of interest - SLIC (Superpixelation Algorithm) and Instance Segmentation using Neural Networks.

The SLIC generates superpixels through clustering pixels based on their colour similarity and proximity in the image plane. Amongst the generated superpixels, clusters belonging to the region of interest are identified through correlating the colour profile of clusters expected to be present in the region of interest. For instance segmentation, the network used is a modified architecture built on the top of UNet. The network is a fully convolutional network with encoder-decoder architecture with multiple skip connections. We have used depthwise separable 2D convolutions instead of standard convolutions as this enables us to get a faster model. In yet another method, we have used dilated convolutions which enables us to reduce the network depth without relying on too many skip connections.\\ \\

\noindent\emph{E.	Feature Extraction}\\

We have extracted colour intensity features across four different colour planes (RGB, CIELab, YCrCb, HSV) and fed them to classification and regression models. Another innovative approach we have employed is using our custom segmentation model for feature extraction also. When one branch of this network is responsible for segmentation, the encoded data is present in the network at an earlier stage as a lower-dimensional vector. Processing this vector through a simple multi-layer perceptron would yield regressed Hb values. This enables us to use transfer learning to train one branch for segmentation outputs. We freeze this part, and use the other branch to get Haemoglobin values.

\end{multicols}

\begin{figure}[h!]
\centering
\includegraphics[scale=0.8]{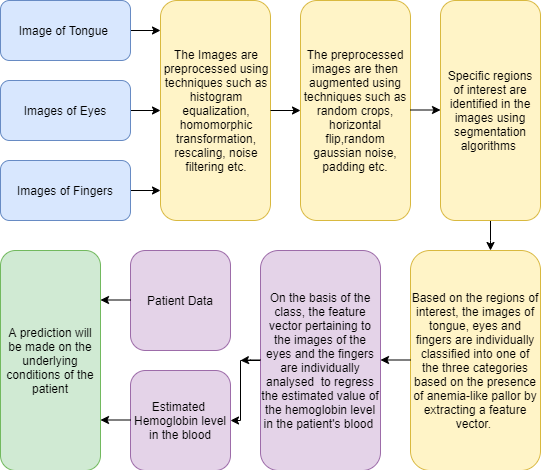}
\caption*{Figure 1 : Pipeline for Haemoglobin Level Detection}
\end{figure}

\begin{multicols}{2}
\noindent\emph{F.	Modelling}\\

We divide our dataset into three categories - severely anaemic, mildly anaemic and non-anaemic. Using classification algorithms and majority rule for the three regions of interest, we make an initial decision regarding the category to which the patient belongs to.  After this, we use regression models specifically trained using samples of each individual category. \\
\\

\noindent\emph{G.	Results}\\

The regression algorithm trained on the three regions of interest gives us the final haemoglobin level of the patient. Table 2 lists haemoglobin levels in different age and gender groups for anaemia diagnosis. With a large dataset, we can use this information for precise prediction. However, at this stage, we are largely ignoring the gender and age of the patient. Taking into account factors like pregnancy, gender and age, a prediction on the cause of the anaemia can also be made.

\end{multicols}

\pagebreak

\begin{figure}[h!]
\centering
\includegraphics[scale=0.45]{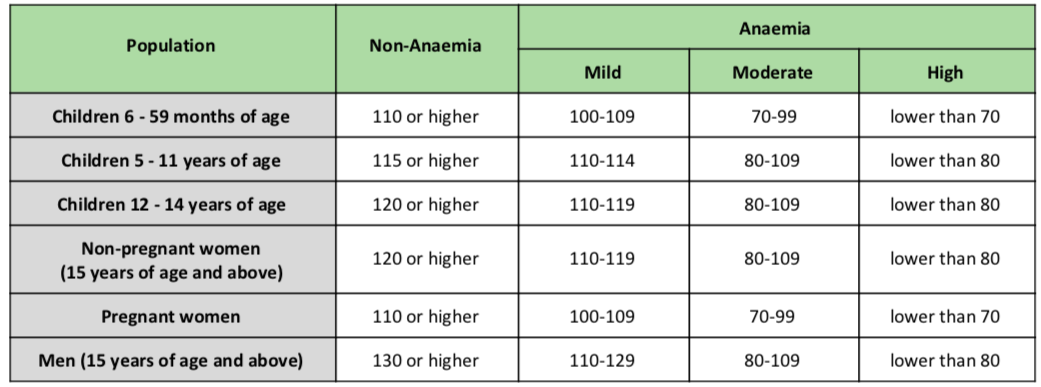}
\caption*{Table 2 :  Haemoglobin levels for anaemia diagnosis
}
\end{figure}

\section{Application Design}
\begin{multicols}{2}
Some subtypes of anaemia require continuous monitoring of the haemoglobin levels in blood. In such cases, deploying the model in an application culminates to immense practical advantage. Personal calibration becomes more convenient, which leads to improved accuracy in the prediction.

Such an application becomes a means for the patient to keep track of his own health. The application also makes it easy to keep records of the gender, age, pregnancy, etc of the patients. With the backend connected to the cloud, the application has vast troves of information being recorded every day. This information can be used to actively and continuously train the prediction model, improving its accuracy over time. Fig 2 shows a picture demonstrating the framework of the application.

\end{multicols}

\section{Further Scope Of Improvement}
\begin{multicols}{2}
\begin{enumerate}
 \item The prediction algorithm can be improved by including patient metadata like age, gender, pregnancy, etc. This can be fed into the algorithm by uploading the patient's blood report along with their conjunctiva images.	
 
 \item The prediction algorithm can be improved in controlled lighting environments with uniform illumination. Certain lighting can also enhance the microvascular structure observed. Neural Networks will be able to pick out small variations in these structures and help study any effect that this has on our previous diagnosis.	
 
 \item The prediction algorithm can be improved by adding a bias term for the altitude at which the test is taken, in the feature vector used for regression of haemoglobin values.
 
 \item The use of a lens will allow for targeted high-resolution pictures of the conjunctiva and also capture variations in microvascular structure.		
 
 \item An additional algorithm can be built to target abnormal pigmentation in the conjunctiva or sclera region. The information thus gathered can be used for the diagnosis of other ailments or help us study the correlation between the ailment and pigmentation observed.
 
 \item A larger global database will give us more opportunity to study haemoglobin level variations and its health impact subject to the geographical location.\\~\\
\end{enumerate}

\end{multicols}

\begin{figure}[h!]
\centering
\includegraphics[scale=0.5]{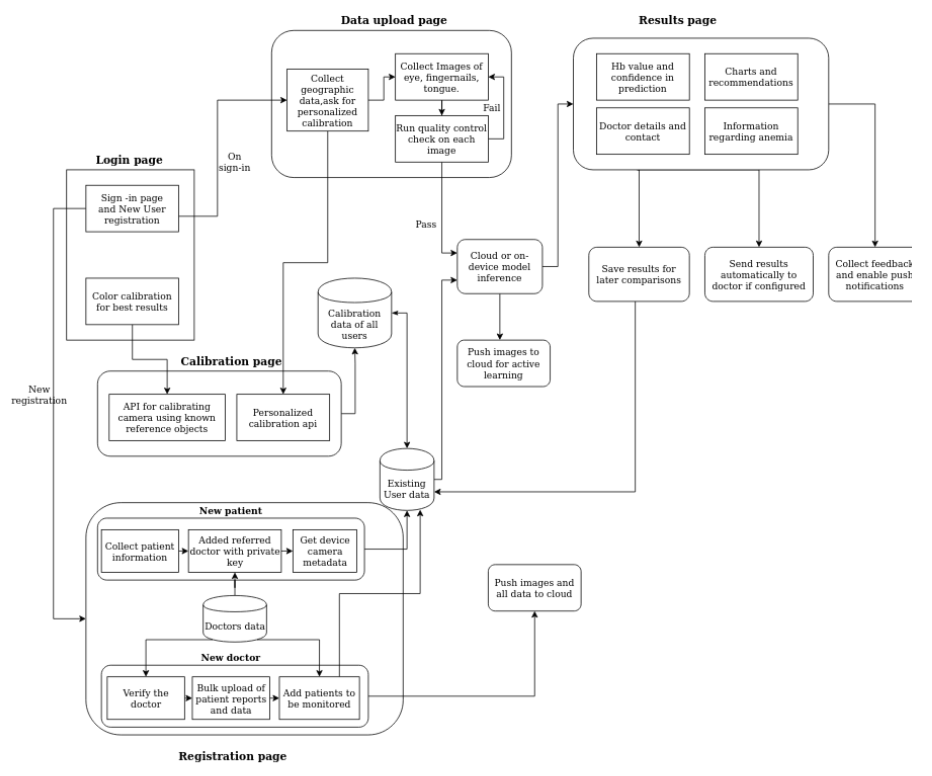}
\caption*{Figure 2 :  Flowchart for Application Design
}
\end{figure}

\pagebreak

\section*{References}
\begin{multicols}{2}

\noindent[1] WHO. Haemoglobin concentrations for the diagnosis of anaemia and assessment of severity. Vitamin and Mineral Nutrition Information System. Geneva, World Health Organization,2011,(WHO/NMH/NHD/MNM/11.1)
(http://www.who.int/vmnis/indicators/haemoglobin. pdf).
					 
\noindent[2] de Benoist B et al., eds. Worldwide prevalence of anaemia 1993-2005. WHO Global Database on Anaemia Geneva, World Health Organization, 2008

\noindent[3] Karnad, A. \& Poskitt, T. The automated complete blood cell count: use of the red blood cell volume distribution width and mean platelet volume in evaluating anemia and thrombocytopenia. Arch. Intern. Med. 145, 1270–1272 (1985).

\noindent[4] Regina D, Sudharshan RC, Rao R. Correlation of pallor with hemoglobin levels and clinical profile of anemia in primary and middle school children of rural Telangana. Int J Contemp Pediatr 2016;3:872-7.

\noindent[5] Sheth, T., BartsSc, B., Choudhry, N., Bowes, M. \& Detsky, A. The relation of conjunctival pallor to the presence of anemia. J. Gen. Intern. Med. 12, 102–106 (1997).

\noindent[6] Robert G. Mannino, David R. Myers, Erika A. Tyburski, Christina Caruso, Jeanne Boudreaux, Traci Leong, G. D. Clifford \& Wilbur A. Lam. Smartphone app for non-invasive detection of anemia using only patient-sourced photos. NATURE COMMUNICATIONS,  (2018) 9:4924 DOI: 10.1038/s41467-018-07262-2
					
\noindent[7] Sohini Roychowdhury and Donny Sun and Matthew Bihis and Johnny Ren and Paul Hage and Humairat H. Rahman. Computer Aided Detection of Anemia-like Pallor. (2017) 1703.05913. (https://arxiv.org/abs/1703.05913)

\noindent[8] John McMurdy, Gregory Jay, Selin Suner \& Gregory Crawford. Photonics-based In Vivo total hemoglobin monitoring and clinical relevance. Published online : 22 April, 2009. Journal of Biophotonics. DOI : 10.1002/jbio.200910019

\noindent[9] Azwad Tamir1, Chowdhury S. Jahan1, Mohammad S. Saif1, Sums U. Zaman1, Md. Mazharul Islam1, Asir Intisar Khan1, Shaikh Anowarul Fattah1, Celia Shahnaz. Detection of Anemia from Image of the Anterior Conjunctiva of the Eye by Image Processing and Thresholding	2017 IEEE Region 10 Humanitarian Technology Conference (R10-HTC) 
DOI: 10.1109/R10-HTC.2017.8289053

\noindent[10] Md. Moin Uddin Atique, Md. Rafiqul Islam Sarker, K Siddique-e-Rabbani. Measurement of Haemoglobin through processing of images of inner eyelid. Bangladesh Journal of Medical Physics. January 2015. DOI : 10.3329/bjmp.v8i1.33929

\noindent[11] Giovanni Dimauro, Danilo Caivano \& Francesco Girardi. A New Method and a Non-Invasive Device to Estimate Anemia Based on Digital Images of the Conjunctiva. IEEE Access SPECIAL SECTION ON HUMAN-CENTERED SMART SYSTEMS AND TECHNOLOGIES.	September 21, 2018. DOI : 10.1109/ACCESS.2018.2867110

\noindent[12] Giovanni Dimauro, Attilo Guarini, Danilo Caivano, Francesco Girardi, Crescenza Pasciolla \& Angela Iacobazzi. Detecting Clinical Signs of Anaemia From Digital Images of the Palpebral Conjunctiva. IEEE Access SPECIAL SECTION ON DATA-ENABLED INTELLIGENCE FOR DIGITAL HEALTH. August 28, 2019. DOI : 10.1109/ACCESS.2019.2932274

\noindent[13] Shaun Collings, Oliver Thompson, Evan Hirst, Louise Goossens, Anup George, Robert Weinkove Non-Invasive Detection of Anaemia Using Digital Photographs of the Conjunctiva. PLOS ONE. DOI:10.1371/journal.pone.0153286 April 12, 2016

\noindent[14] Tadaki Kawanabe, Meiko Tanigawa, Sachiko Kazizaki, Nur Diyana Kamarudin, Xiaoyu Mi, Toshihiko Hanawa \& Hiroshi Odaguchi. Correlation between tongue body colour, as quantified by machine learning, and clinical indices. Advances in Integrative Medicine 7 (2020). Elesevier Feb 18, 2019.

\end{multicols}

\end{document}